\documentclass[runningheads]{llncs}

\usepackage[T1]{fontenc}
\usepackage{bbding, amsmath, booktabs, array, epsfig, xspace, todonotes, soul, multirow, graphicx, booktabs, pifont, wasysym, amssymb, makecell, threeparttable}
\usepackage[hidelinks]{hyperref}
\begin{document}
\title{GSTran: Joint Geometric and Semantic Coherence for Point Cloud Segmentation}

\author{\small Abiao Li\inst{1} \and
        Chenlei Lv\inst{2} \and
        Guofeng Mei\inst{3} \and
        Yifan Zuo \inst{1} \and
        Jian Zhang \inst{4} \and
        Yuming Fang\inst{1}\thanks{Corresponding Author-Yuming Fang}}

\institute{Jiangxi University of Finance and Economics, NanChang 330013, China \\
           \email{\{lab183,leo.fangyuming\}@foxmail.com, kenny0410@126.com}\and 
           Shenzhen University, ShenZhen 518060, China \\
           \email{chenleilv@mail.bnu.edu.cn} \and
           Fondazione Bruno Kessler, Trento 38123, Italy \\
            \email{gfmeiwhu@outlook.com} \and
           University of Technology Sydney, Sydney 2007, Australia\\
           \email{jian.zhang@uts.edu.au}}

\maketitle
\begin{abstract}
Learning meaningful local and global information remains a challenge in point cloud segmentation tasks. When utilizing local information, prior studies indiscriminately aggregates neighbor information from different classes to update query points, potentially compromising the distinctive feature of query points. In parallel, inaccurate modeling of long-distance contextual dependencies when utilizing global information can also impact model performance. To address these issues, we propose GSTran, a novel transformer network tailored for the segmentation task.  The proposed network mainly consists of two principal components: a local geometric transformer and a global semantic transformer. In the local geometric transformer module, we explicitly calculate the geometric disparity within the local region. This enables amplifying the affinity with geometrically similar neighbor points while suppressing the association with other neighbors. In the global semantic transformer module, we design a multi-head voting strategy. This strategy evaluates semantic similarity across the entire spatial range, facilitating the precise capture of contextual dependencies. Experiments on ShapeNetPart and S3DIS benchmarks demonstrate the effectiveness of the proposed method, showing its superiority over other algorithms.
The code is available at \href{https://github.com/LAB123-tech/GSTran}{https://github.com/LAB123-tech/GSTran.}

\end{abstract}

\section{Introduction}
3D point cloud segmentation is a crucial topic in computer vision and graphics, with widespread applications in autonomous driving, simultaneous localization and mapping (SLAM), augmented reality, and virtual reality. Limited by the weak discriminative capability of handcraft features, traditional solutions fail to achieve satisfactory segmentation performance. Fortunately, benefited from the development of deep learning, the performance of segmentation has been significantly improved to support high-level semantic analysis~\cite{mei2024unsupervised}.

Recent efforts~\cite{1-pointnet++,3-MultiScaleAttention} have shown promising results in point cloud processing by integrating multi-scale neighbor features to enhance the capability of feature analysis. However, repeating similar contextual features at various scales seems redundant and computationally expensive, especially for hierarchical architectures. Subsequent studies~\cite{4_GBNet,5-GNN,6-SCConv}, advanced the refinement of point features by encoding the detailed geometric description information in the local region of the point cloud. Nonetheless, these algorithms simply stack the geometric and coordinate information of the point cloud and take the stacked result as input to the network. These approaches fail to fully exploit geometric properties, as geometric information is likely to be lost during learning. Large language model (LLM) based methods~\cite{7-partslip,8-ponderv2} have also demonstrated their effectiveness in point clouds. The main idea of these approaches is to render the 3D point cloud as a set of multi-view 2D images for semantic parsing. But, LLM, being a type of generalized model, lacks the ability to perceive the internal structure of point clouds~\cite{mei2024geometrically}. For accurate geometric structure determination of point clouds, the LLM still have limitations.

With the breakthrough of transformer in the fields of natural language processing and computer vision, some algorithms~\cite{9_PoinTr,10-PointGT} consider incorporating geometric information of the point cloud into the self-attention mechanism to execute segmentation. PointTr~\cite{9_PoinTr} introduces a geometric-aware module that models the local geometric relationships of point clouds by constructing neighborhood graph structures. PointGT~\cite{10-PointGT} decouples the local neighborhoods and utilizes a bi-directional cross-attention mechanism to merge the edge and inside components of the local features. Nonetheless, the weights assigned to neighbor points rely solely on learning. The model struggles to prioritize neighbor points belonging to the same category as the query point. As a result, the descriptive power of point features is compromised, potentially leading to erroneous segmentation results at the boundary.

Although deploying transformer in the local regions of point clouds plays a positive role in capturing geometric information, its receptive field remains limited. In point clouds, points that are semantically related to query points may be located far apart. Therefore, certain research studies~\cite{11-GTNet,12-3DCTN,13-SelfPosition} employ transformer to compute global similarity for each point. Such approaches treat all points as neighbor points of the query point. They employ self-attention to capture long-range contextual dependencies among points of the same class. This contributes to the enhancement of semantic understanding. However, the strong similarity between points computed with self-attention does not guarantee that these points belong to the same category. 

In light of the above analysis, fully exploiting local geometric features and accurately capturing long-range dependencies hold significant importance for transformer to understand point clouds. Based on this, we propose a novel transformer architecture for point cloud segmentation, named GSTran. It consists of two crucial transformer modules: (i) a local geometric transformer and (ii) a global semantic transformer. In the local geometric transformer, we thoroughly investigate the geometric disparity in the local region to quantify the significance of each neighbor point. Specially, we explicitly compute the distance from the query point to the tangent plane of its corresponding neighbor points. In general, the query point tends to be closer to the tangent plane of its geometrically similar neighbors. As a result, greater significance is attributed to these neighbor points, leading to enhanced segmentation outcomes for boundaries within the point cloud.

In the global semantic transformer, a multi-head voting strategy is introduced to facilitate the preservation of meaningful long-range dependencies. Essentially, this involves employing a multi-head attention mechanism to compute multiple global similarities for each point. However, unlike multi-head attention mechanisms, we further extract the shared information from multiple global similarities to generate a global mask. By leveraging the global mask, we can more accurately compute the long-distance similarity between points in the point cloud. Equipped with both transformer modules, GSTran effectively infuses features with rich local geometric structures and comprehensive global semantic context. The main contributions of our paper are summarized as follows:

\begin{itemize}
    \item We introduce a local geometric transformer module that leverages the geometric disparity within the local point cloud. This module assigns higher weights to neighbor points that exhibit similar geometric structure to the query point. This enhancement improves the model's ability to distinguish target boundaries.
    \item We design a multi-head voting-based global semantic transformer module to capture semantically aligned contextual dependencies beyond local regions. By leveraging this module, we can enhance the accuracy of computing long-distance similarity between points within the point cloud.
    \item We present the performance evaluation of GSTran on both ShapeNetPart and S3DIS benchmarks to demonstrate the efficacy of our approach in addressing segmentation tasks. 
\end{itemize}

\begin{figure}[t]
  \centering
  \includegraphics[width=1.0\linewidth]{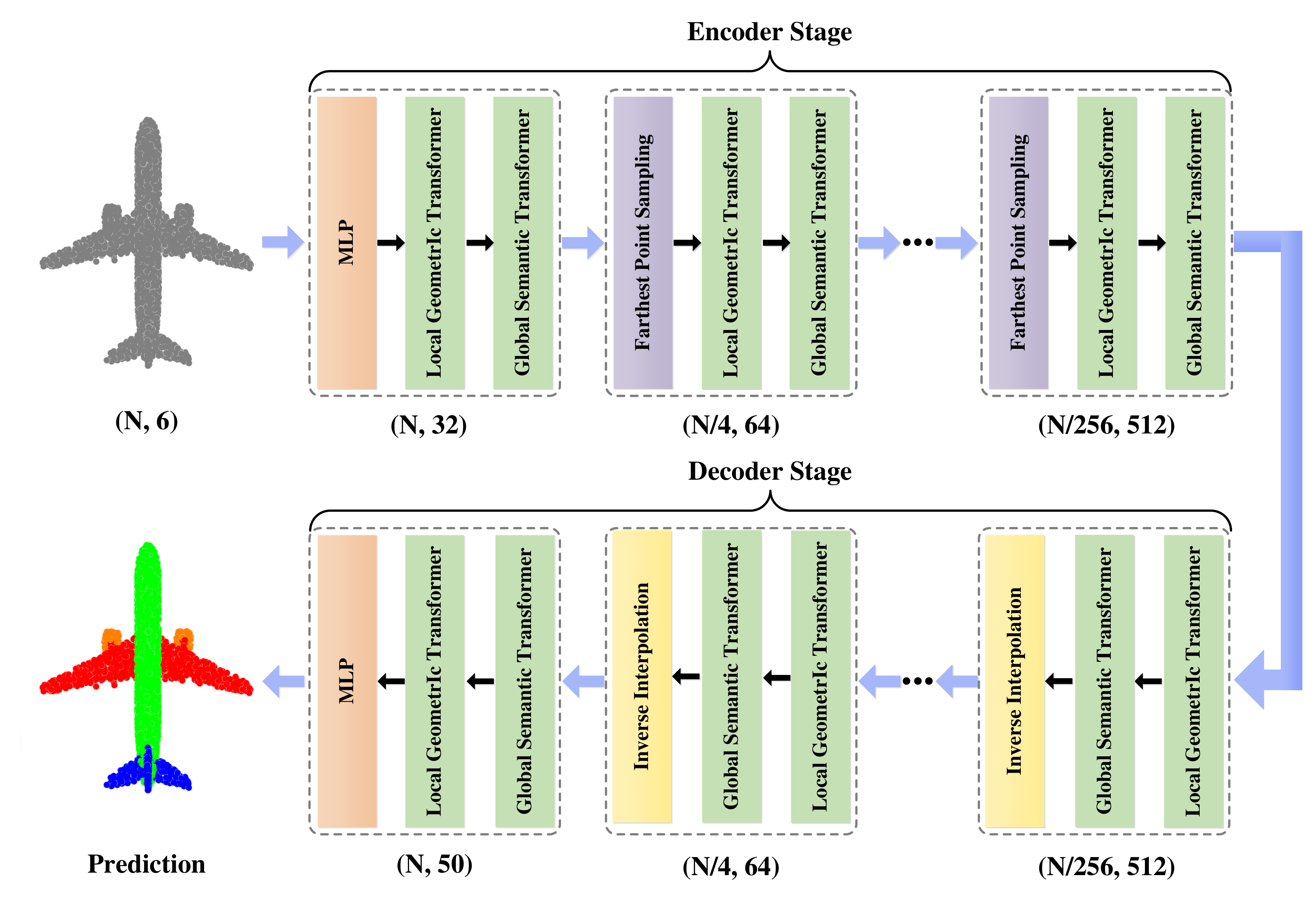}
  \caption{Overview of the proposed model. In the encoder and decoder stages, the transformer structure serves as the primary feature aggregator throughout the network. MLP: Multi-layer perception. N: The number of points in the point cloud.}
  \label{f1}
\end{figure}

\section{Methodology}
\subsection{Overview}
The architecture of the proposed model, as shown in Fig.~\ref{f1}, features a hierarchical framework consisting of symmetric encoder and decoder stages. There are five stages in the encoder. The point cloud sampling rate between two adjacent stages is 1/4, and the channel expansion rate is 2. Within the first stage of the encoder, the MLP layer projects the point cloud data with coordinate and normal vector information into higher dimensional feature. Subsequently, two sequential modules – local geometric and global semantic transformer – are sequentially applied for feature extraction at different scales. The features are progressively downsampled with channel expanding in the following four stages. This process continues until the sampling rate reaches 1/256 and the channel expands to 512. The decoder stage follows a similar structure but utilizes inverse interpolation~\cite{1-pointnet++} for progressive upsampling. Details about the local geometric and global semantic transformer modules are given in Section 2.2 and Section 2.3, respectively.

\subsection{Local Geometric Transformer Module}
\begin{figure}[t]
  \centering
  \includegraphics[width=1.0\linewidth]{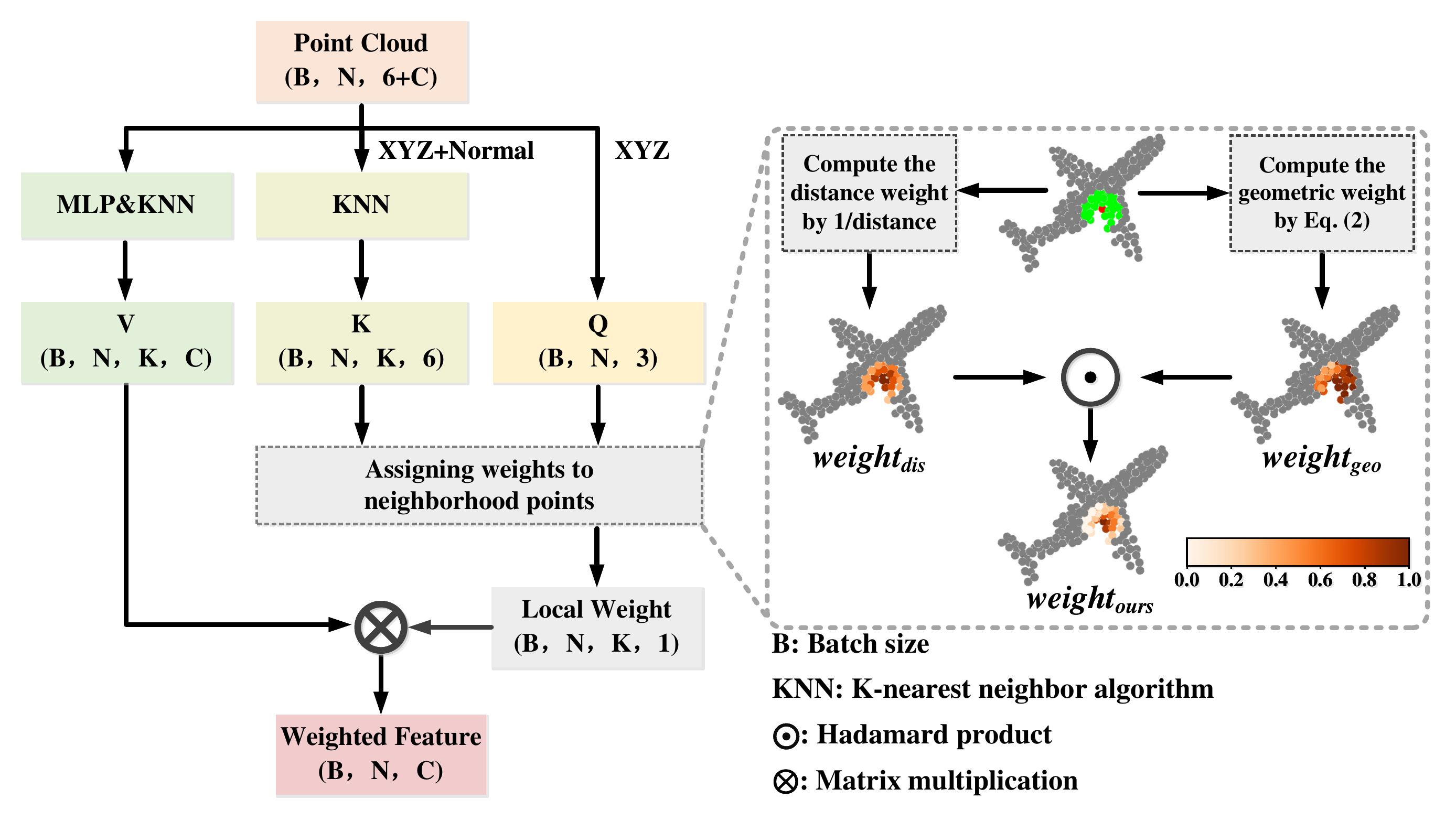}
  \caption{Structure of the local geometric transformer module. We visualize the local weight on an airplane, with a red query point located on the wing. In \textit{weight$_{ours}$}, the weights of neighbor points in the wing decay slowly as the distance increases. However, the weights of other neighbor points decay rapidly.}
  \label{f2}
\end{figure}

The local geometric transformer module is designed to achieve discriminative feature extraction. It achieves this by investigating the geometric disparity within the local region. As depicted in Fig.~\ref{f2}, given input point cloud $\chi=\{x_{i}|i = 1, 2,\cdots N\} \in \mathcal R^{N\times (6+C)}$, each point $x_{i}$ is defined by its position coordinate $p_i\in \mathcal R^3$, normal vector $n_{i} \in \mathcal R^3$ and feature $f_i \in \mathcal R^C$. We separate the coordinate information as the $Q$ vector. At the same time, corresponding neighbor sets are constructed for each point in both Euclidean space and feature space using the KNN algorithm, represented by $K$ and $V$, respectively. Subsequently, corresponding weights are assigned to the neighbor points to quantify their importance with respect to the query points. 

As shown in the dashed box of Fig.~\ref{f2}, given a red query point $p_i$ (located at the wing) and its green neighbor points $h=\{q_{1}, \cdots, q_{k}\} \in \mathcal R^{k\times3}$, we compute the reciprocal of the Euclidean distance to obtain the distance weights denoted as \textit{weight$_{dis}$}. Clearly, smaller weights are assigned to distant neighbor points, but the weights remain consistent for neighbor points of different classes, such as fuselage and wing. With only distance weights, the query point may indiscriminately aggregate information from neighbor points of various classes. This may compromise the descriptive power of the query point.

\begin{figure}[t]
  \centering
  \includegraphics[width=\linewidth]{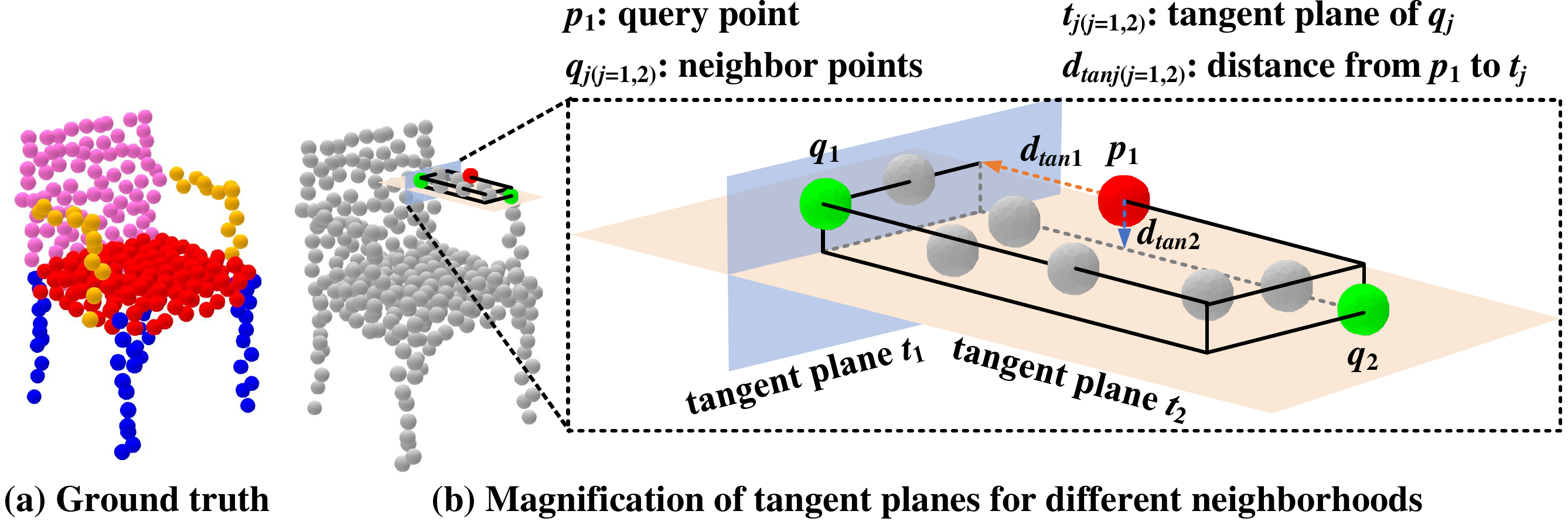}
  \caption{Illustration of the distance $d_{tan1}$ and $d_{tan2}$ from $p_1$ to the tangent plane of $q_{1}$ and $q_{2}$, respectively. Although the Euclidean distance from $p_1$ to both $q_1$ and $q_2$ remain the same, $p_1$ is closer to $t_2$, signifying that $q_2$ holds greater significance than $q_1$.}
  \label{f3}
\end{figure}

Therefore, we use the following formula~\cite{14_PointSample} to characterize the geometric importance of neighbor points with respect to the query point.
\begin{equation}
d_{tan}=(p_{i}-q_{j})n_{j}, \; \forall j \in G_{i},
\label{e1}
\end{equation}
where $p_{i}$ is the coordinate of \textit{i}-th query point, $q_{j}$ denotes the coordinate of \textit{j}-th neighbor point corresponding to $p_{i}$. $G_{i}$ is the index set of the local group centered on $p_{i}$. $n_{j}$ is the normal vector corresponding to $q_{j}$. Essentially, $d_{tan}$ signifies the distance from the query point to the tangent plane of the neighbor points, as shown in Fig.~\ref{f3}. (Note: to clearly explain $d_{tan}$, we choose a chair target for illustration.) The tangent plane of neighbor points belonging to the same class as the query point is close to the query point. This proximity implies that the importance attached to this neighbor point should be large. Based on $d_{tan}$, we quantify the weight in the geometry using an exponential function as follows.

\begin{equation}
weight_{geo}=exp(-d_{tan})
\label{e2}
\end{equation}
Eq.~\eqref{e2} shows that the weight is inversely proportional to the value of \textit{$d_{tan}$}. It can be seen from Fig.~\ref{f2} that the \textit{{weight}$_{geo}$} assigns higher weights to points on the wings and lower weights to points on the fuselage. However, the weights assigned to the neighbor points from the wings are nearly equal in geometric weight. Following the principle that larger distances correspond to fewer dependencies between points, we consider combining the distance and geometric weights through a Hadamard product to obtain the local weight of ours. It can be observed that the weights of neighbor points on the wing in \textit{weight$_{ours}$} exhibit a slow decay rate with increasing distance. This is in contrast to points located on the fuselage, where the weights decrease more rapidly. High response weight values are predominantly distributed in the wings. Finally, the weighted features are obtained by multiplying the local weight with the $V$ vector.

\begin{figure}[ht]
\vspace{-.2cm}
  \centering
  \includegraphics[width=1.0\linewidth]{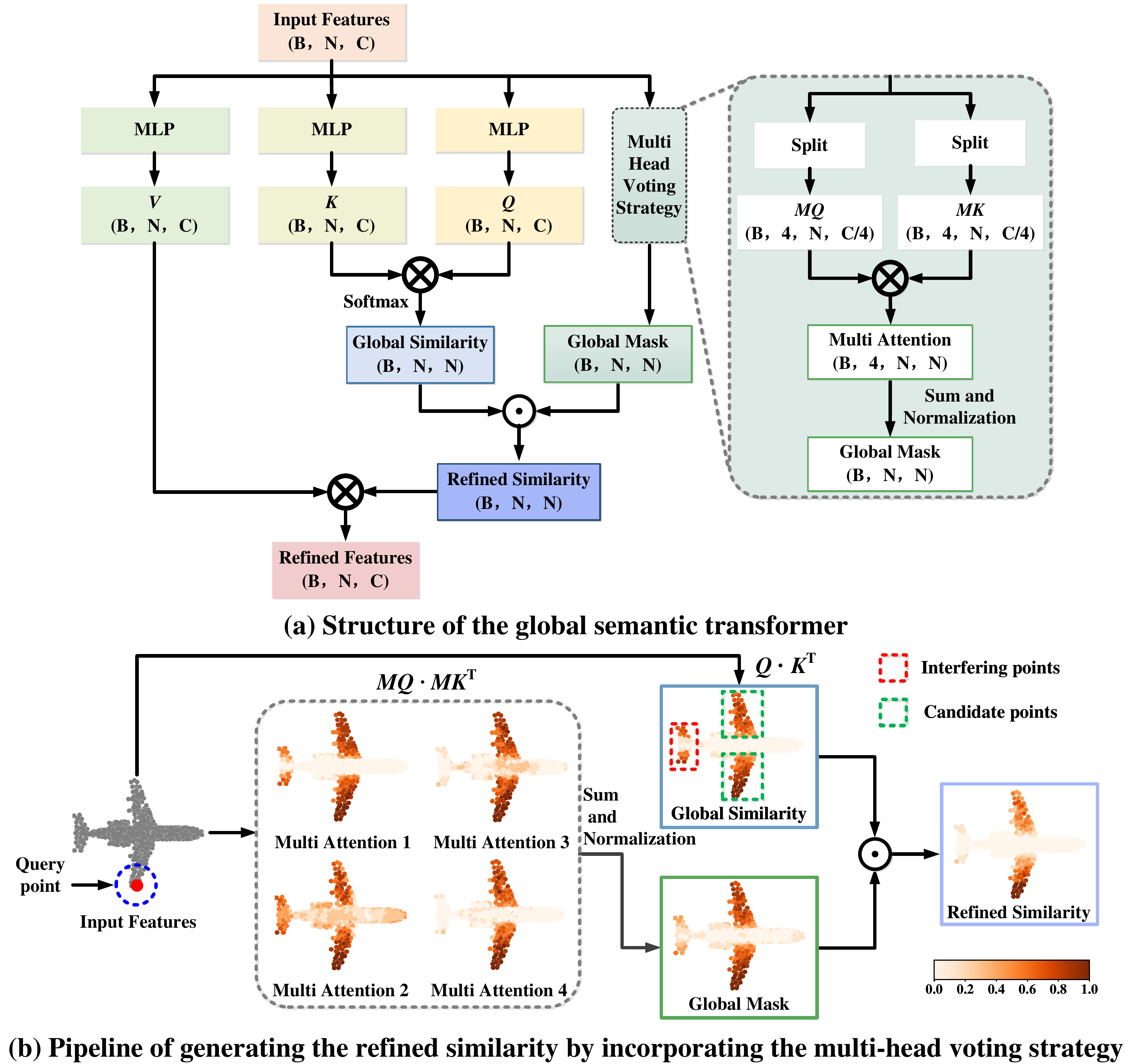}
  \caption{Overview of the global semantic transformer module. We visualize the refined similarity on an airplane, with a red query point located at the wing. In the refined similarity generated by our method, high response weights are exclusively assigned to points belonging to the wing section.}
  \label{f4}
  \vspace{-.2cm}
\end{figure}

\subsection{Global Semantic Transformer Module}
The output of the local geometric transformer module exhibits powerful discriminative capabilities for local features. However, the relations captured by this module are restricted to local structures within the point cloud. Certain approaches~\cite{11-GTNet,12-3DCTN,15-MCTNet} employ transformer to calculate the global similarity between each pair of points in the entire point cloud. Then, the global similarity, along with the point cloud features, is multiplied to perform feature aggregation. Mathematically, the formula is as follows:
\begin{equation}
y_i=\sum_{f_j\in \chi}\text{softmax}\left(\frac{Q K^{\text{T}}}{\sqrt{d_k}}\right)V, \quad Q=\varphi(f_i), K=\psi(f_j), V=\phi(f_j)
\label{e3}
\end{equation}
where $f_{i}$ denotes the feature of the \textit{i}-th query point, and $f_{j}$ represents the feature of the \textit{j}-th point within point cloud $\chi$. The symbols $\varphi$, $\psi$, and $\phi$ represent three MLP operations conducted on the feature of the point clouds to obtain the query $Q$, key $K$, and value $V$ vectors, respectively. $d_k$  is the dimension of the feature. 

A drawback of using $QK^{\text{T}}$ in Eq.~\eqref{e3} to compute global similarity is that the obtained global similarity may not accurate. To be specific, points that show strong similarity to query points may belong to different classes. A example of the global similarity is shown in Fig.~\ref{f4}(b). Given a query point within the wing section (marked in red), we visualize the global similarity corresponding to this query point. It is evident that the global similarity exhibits high response weights for points in both the wing and tail sections. However, the weights of the points in the wing section are the only ones truly relevant. We designate points in the tail as interfering points and points in the wings as candidate points.

To accurately compute the long-distance similarity between points within the point cloud, we devise a multi-head voting strategy that is incorporated into the global semantic transformer module. The corresponding structure is illustrated in Fig.~\ref{f4}. Given the features of the input point cloud, three MLP operations are performed to obtain the $Q$, $K$, and $V$ vectors, respectively. Then, the matrix multiplication between $Q$ and the transposition of $K$ is performed, followed by the softmax function to generate the global similarity for each point. In order to mitigate the effect of interfering points, we further refine the global similarity by generating a global mask using a multi-head voting strategy. The details are depicted within the dashed box in Fig.~\ref{f4}(a). 

Specifically, we split the channels of the point cloud to obtain multi-head representations of the feature, denoted as $MQ$. Similarly, we do the same to obtain $MK$. We proceed by performing matrix multiplication between $MQ$ and the transpose of $MK$ to produce multiple attentions. Corresponding results are shown in the dashed box in Fig. 4(b). It can be observed that the weights of candidate points consistently exhibit strong response in multi-attention 1 to 4. However, the weights of the interfering points exhibit randomness. For instance, in multi-attention 1 and 2, the weights of the interfering points are large, whereas in multi-attention 3 and 4, they are small. This inconsistency leads to uncertainty in selecting the optimal attention. To mitigate the influence of randomness, we propose to summarize multiple attentions and apply normalization to obtain a global mask. It can be observed that in the global mask, the weights of the interfering points are reduced to some extent. To further reduce the weights of interfering points, we consider subjecting the global similarity and global mask to Hadamard product operations to generate refined similarity. In the refined similarity, the weights of the interfering points are significantly reduced, while the weights of the candidate points still exhibit strong responses. Finally, the refined similarity is multiplied with the vector $V$ to obtain the refined features, which fulfill the aggregation of global information for each point.

\section{Experiments}
\subsection{Experimental setting}
We demonstrate the effectiveness of the proposed model in various point cloud segmentation tasks. Specifically, we utilize the S3DIS dataset for 3D semantic
segmentation and the ShapeNetPart dataset for 3D part segmentation.  Experiments were conducted on an Ubuntu system equipped with two NVIDIA GTX 2080Ti GPUs. We employed the Adam optimizer with momentum and weight decays set to 0.9 and 0.0001, respectively.  For the S3DIS dataset, we trained for 60,000 iterations, starting from an initial learning rate of 0.5. This rate drops by a factor of 10 at steps of 30,000 and 50,000.  For the ShapeNetPart dataset, we trained for 200 epochs. The initial learning rate is set to 0.05 and reduced by a factor of 10 at epochs 100 and 150.

\subsection{3D semantic segmentation and part segmentation}

\begin{table}[t]
\caption{Semantic segmentation results on S3DIS dataset.}
\label{t1}
\tabcolsep 4pt
\resizebox{\columnwidth}{!}{
\begin{threeparttable}
\centering
\begin{tabular}{l*{16}{c}}
\toprule[1.3pt]
Method & mIoU & mAcc & OA & ceiling & floor & wall & beam & column & window & door & table & chair & sofa & bookcase & board & clutter \\
\midrule[0.5pt]
PointWeb~\cite{17_PointWeb} & 66.7 & 76.2 & 87.3 & 93.5 & 94.2 & 80.8 & 52.4 & 41.3 & 64.9 & 68.1 & 71.4 & 67.0 & 50.3 & 62.7 & 62.2 & 58.5 \\
PointAttn.~\cite{18_PointAttention} & 66.3 & 77.3 & 88.9 & 94.3 & 97.0 & 76.0 & 64.7 & 53.7 & 59.2 & 58.8 & 72.4 & 69.2 & 42.6 & 60.8 & 54.1 & 59.0 \\
SCFNet~\cite{19-SCFNet}  & 71.6 & 82.7 & 88.4 & 93.3 & 96.4 & 80.9 & 64.9 & 47.4 & 64.5 & 70.1 & 81.6 & 71.4 & 64.4 & 67.2 & \textbf{67.5} & 60.9 \\
CBL~\cite{20_CBL} & 73.1 & 79.4 & 89.6 & 94.1 & 94.2 & 85.5 & 50.4 & \textbf{58.8} & 70.3 & \textbf{78.3} & 75.7 & 75.0 & 71.8 & 74.0 & 60.0 & 62.4 \\
PointTrans.~\cite{21-PointTransformer} & 73.5 & 81.9 & 90.2 & 94.3 & 97.5 & 84.7 & 55.6 & 58.1 & 66.1 & 78.2 & 77.6 & 74.1 & 67.3 & 71.2 & 65.7 & \textbf{64.8} \\
BAAFNet~\cite{22_BAAFNet} & 72.2 & 83.1 & 88.9 & 93.3 & 96.8 & 81.6 & 61.9 & 49.5 & 65.4 & 73.3 & 72.0 & 83.7 & 67.5 & 64.3 & 67.0 & 62.4 \\
DPFA ~\cite{29-DPFA_Net} & 61.7 & 61.6 & 89.2 & \textbf{94.6} & \textbf{98.0} & 79.2 & 40.7 & 36.6 & 52.2 & 70.8 & 65.9 & 74.7 & 27.7 & 49.8 & 51.6 & 60.6 \\
RepSurf~\cite{23-RepSurf-U} & 74.1 & 82.6 & 90.8 & 93.8 & 96.3 & \textbf{85.6} & 62.5 & 52.5 & 67.4 & 75.3 & 73.9 & 82.1 & 71..5 & 73.3 & 65.1 & 61.8 \\
Ours & \textbf{74.9} & \textbf{83.5} & \textbf{91.3} & 93.2 & 96.1 & 85.1 & \textbf{65.1} & 50.7 & \textbf{71.2} & 73.3 & \textbf{79.1} & \textbf{84.2} & \textbf{71.9} & \textbf{73.9} & \textbf{67.5} & 62.4 \\ 
\bottomrule[1.3pt]
\end{tabular}
\begin{tablenotes}[flushleft]
\small
\item Note: bold font indicates best result.
\end{tablenotes}
\end{threeparttable}}
\end{table}

\begin{table*}[ht]
\centering
\tabcolsep 4pt
\caption{Part segmentation results on ShapeNetPart dataset.}
\label{t2}
\resizebox{\textwidth}{!}{
\begin{threeparttable}
\begin{tabular}{@{}>{\raggedright}p{2cm}ccccccccccccccccccc@{}}
\toprule[1.3pt]
Methods & \begin{tabular}[c]{@{}c@{}}Ins.\\ mIoU\end{tabular} & \begin{tabular}[c]{@{}c@{}}Cat.\\ mIoU\end{tabular} & air. & bag & cap & car & cha. & ear. & gui. & kni. & lam. & lap. & mot. & mug & pis. & roc. & ska. & tab. \\
\midrule[0.5pt]
PointAttn.~\cite{18_PointAttention} & 85.9 & 84.1 & 83.3 & 86.1 & 85.7 & 80.3 & 90.5 & 82.7 & 91.5 & 88.1 & 85.5 & 95.9 & 77.9 & 95.1 & 84.0 & 64.3 & 77.6 & 82.8 \\
PointASNL\cite{24_PointASNL} & 86.1 & 83.4 & 84.1 & 84.7 & 87.9 & 79.7 & 92.2 & 73.7 & 91.0 & 87.2 & 84.2 & 95.8 & 74.4 & 95.2 & 81.0 & 63.0 & 76.3 & 83.2 \\
PointMLP~\cite{25_PointMLP} & 86.1 & 84.6 & 83.5 & 83.4 & 87.5 & 80.5 & 90.3 & 78.2 & 92.2 & 88.1 & 82.6 & 96.2 & 77.5 & \textbf{95.8} & 85.4 & \textbf{64.6} & \textbf{83.3} & 84.3 \\
APES~\cite{26-APES} & 85.8 & 83.9 & 85.3 & 85.8 & 88.1 & \textbf{81.2} & 90.6 & 74.0 & 90.4 & \textbf{88.7} & 85.1 & 95.8 & 76.1 & 94.2 & 83.1 & 61.1 & 79.3 & 84.2 \\
PointTran.~\cite{21-PointTransformer} & 86.5 & 83.7 & 85.8 & 85.3 & 86.8 & 77.2 & 90.5 & 82.0 & 90.8 & 87.5 & 85.2 & 96.3 & 75.4 & 93.5 & 83.8 & 59.7 & 77.5 & 82.5 \\
PointGT~\cite{27_PointGT} & 85.8 & 84.2 & 84.3 & 84.5 & 88.3 & 80.9 & 91.4 & 78.1 & 92.1 & 88.5 & 85.3 & 95.9 & 77.1 & 95.1 & 84.7 & 63.3 & 75.6 & 81.4 \\
PCT~\cite{30-PCT} & 86.4 & 83.1 & 85.0 & 82.4 & 89.0 & \textbf{81.2} & 91.9 & 71.5 & 91.3 & 88.1 & 86.3 & 95.8 & 64.6 & 95.8 & 83.6 & 62.2 & 77.6 & 83.7 \\
LGGCM~\cite{28_LGGCM} & 86.7 & 84.8 & 85.1 & 85.9 & \textbf{90.3} & 80.8 & 91.6 & 75.4 & \textbf{92.7} & 88.1 & 86.5 & 96.1 & 77.0 & 94.2 & 84.5 & 63.6 & 80.2 & 84.3 \\
Ours & \textbf{87.5} & \textbf{85.6} & \textbf{86.1} & \textbf{87.2} & 88.1 & 79.4 & \textbf{92.4} & \textbf{82.3} & 92.0 & 88.4 & \textbf{86.9} & \textbf{96.7} & \textbf{78.7} & 95.6 & \textbf{85.8} & 63.8 & 79.3 & \textbf{85.3} \\
\bottomrule[1.3pt]
\end{tabular}
\begin{tablenotes}[flushleft]
\small
\item Note: air.: airplane. cha.: chair. ear.: ear-phone. gui.: guitar. kni.: knife. lam.: lamp. lap.: laptop. mot.: motorbike. pis.: pistol. roc.: rocket.
\item \hspace{9.0mm}ska.: skateboard. tab.: table.
\end{tablenotes}
\end{threeparttable}}
\end{table*}

\noindent\textbf{Semantic Segmentation.} The S3DIS dataset comprises 271 scenes from 6 indoor areas, and each point labeled among 13 categories. Since the S3DIS dataset does not provide normal vector information, it becomes necessary for us to compute the normal vector before training. In this step, we refer to the method outlined in \cite{16_NormalEstimate}. Here, we extract the eigenvector corresponding to the minimum eigenvalue of the point cloud. This is achieved by performing singular value decomposition on the covariance matrix of the point cloud. At the same time, we can precompute the normal vector for each point before training, storing them in the dataset to reduce running time.

During the training process, the computational complexity of the global semantic transformer module being $O(n^2)$, where $n$ represents the number of points processed. However, in the entire algorithm process, only the first global transformer module in the encoding stage and the last one in the decoder stage handle a relatively large number of points. The other transformer modules, due to the UNet~\cite{31-UNET} structure used in the algorithm, process a smaller number of points. Moreover, we use block-wise training strategy~\cite{1-pointnet++} to ensure that $n$ is not excessively large, thereby improving running efficiency. Specially, each room is divided into 2m$\times$2m blocks, from which 4096 points are randomly sampled for training.  During testing, we employ six-fold cross-validation for evaluation. This approach involves using all points in the scene for testing purposes. For evaluation metrics, we use mean instance IoU (mIoU), mean class accuracy (mAcc), and overall pointwise accuracy (OA).

\begin{figure*}[ht]
  \centering
  \includegraphics[width=\linewidth]{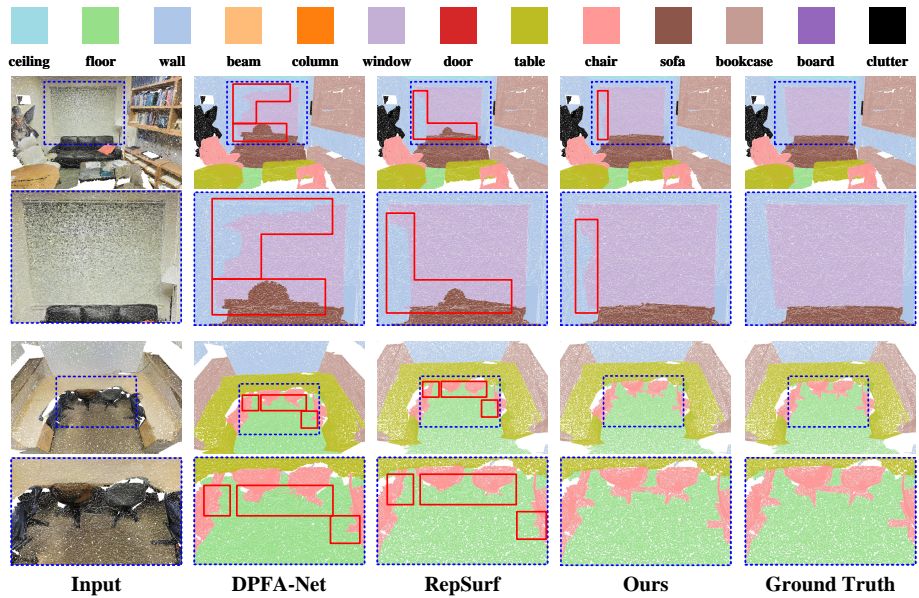}
  \caption{Visualization of segmentation results for different methods on S3DIS. The red box indicates the region where the segmentation error occurs.}
  \label{f5}
\end{figure*}

As shown in Table \ref{t1}, compared with other algorithms, the proposed algorithm achieves satisfied results in mIoU, mAcc, and OA. The superior performance over other transformer architecture~\cite{21-PointTransformer} (+1.4\% mIoU, 1.6\% mAcc, 1.1\% OA) proves the importance of incorporating the geometric relationships in semantic segmentation. Meanwhile, we compare the segmentation results of our algorithm with several algorithms in various scenarios on the S3DIS dataset. The corresponding results are shown in Fig.~\ref{f5}. In scene one (top row), compared to other algorithms, our proposed algorithm does not show significant errors in the sofa segmentation results. Due to the fact that windows are nearly embedded within walls, both our method and other algorithms suffer from certain shortcomings in the segmentation results of windows. But, the segmentation results of our method for the window are the closest to the ground truth. In scenario two (third row), our algorithm effectively improves the segmentation results of chairs, especially at the junction areas with the floor.


\textbf{Part Segmentation.} The ShapeNetPart dataset consists of 16, 880 3D models categorized into 16 shape categories with 50 different parts. For each point cloud object, 2048 points are uniformly sampled from its surface for both the training and testing phases. The data augmentation strategy we applied involves random scaling of the object within a range of 0.8 to 1.25, coupled with arbitrary rotation along any coordinate axis. We evaluate the performance of our method using Intersection over Union (IoU) for each category. Additionally, we calculate the average IoU of all instances (Ins.mIoU) and the average IoU of all categories (Cat.mIoU) for the entire dataset.

\begin{figure}[!ht]
    \centering
    \begin{minipage}[t]{\linewidth}
        \centering
        \includegraphics[width=0.97\linewidth]{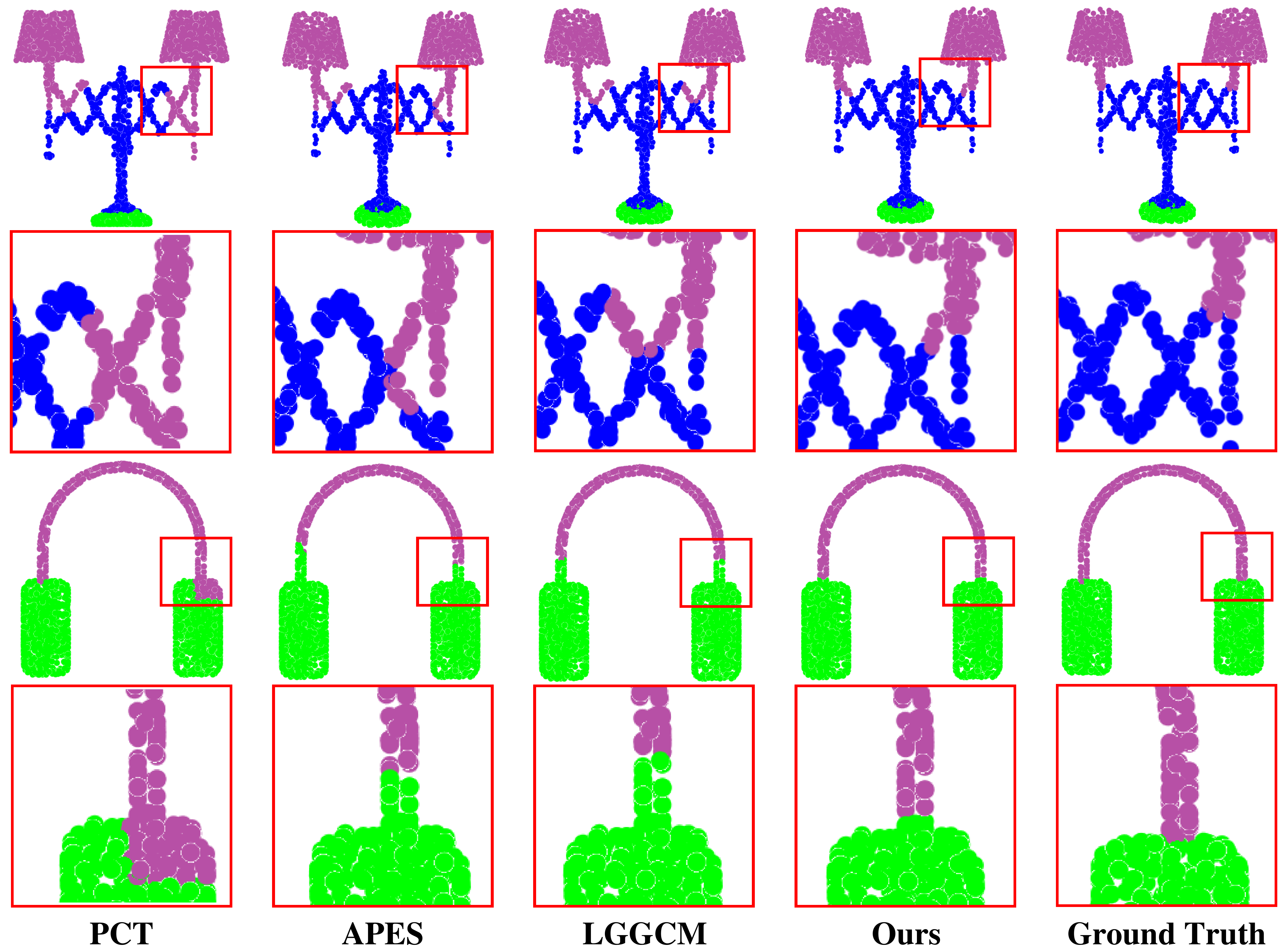}
        \caption{Visualization of segmentation results across various methods on ShapeNetPart, with areas of inaccurate segmentation highlighted by a red box.}
        \label{f6}
    \end{minipage}
    
    \begin{minipage}[t]{\linewidth}
        \centering
        \includegraphics[width=0.97\linewidth]{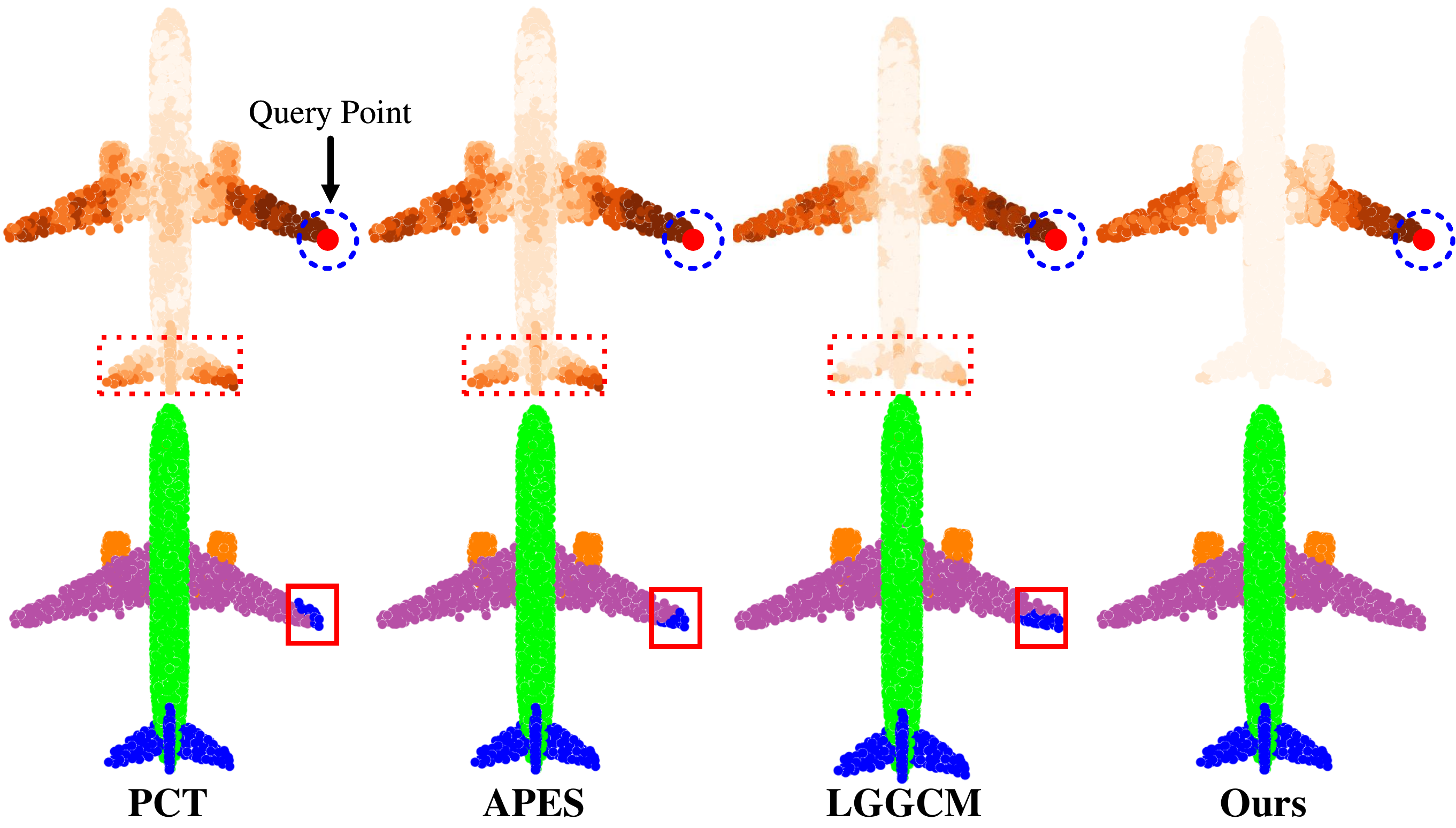}
        \caption{Visualization of the global similarity and the corresponding segmentation results from various methods. The red dashed box indicates interfering points with high response weights. The red solid line box indicates the incorrect segmentation result.}
        \label{f7}
    \end{minipage}
    \vspace{-.2cm}
\end{figure}

Table~\ref{t2} presents the performance comparison of various algorithms on the ShapeNetPart dataset. Although the performance is quite saturated, our method achieves the best performance as measured by Ins.mIoU and Cat.mIoU. Compared to PCT~\cite{30-PCT}, which also employs the concept of transformer, our algorithm achieves 1.1\% improvement in Ins.mIoU. In addition, it achieved a 2.5\% improvement in cat.mIoU, further demonstrating the effectiveness of our approach. Meanwhile, our method achieves the best performance on 9 out of 16 categories, such as airplane, bag and chair. Since the geometric discrepancies exhibited in the local regions of these categories are distinct. This property facilitates the local geometric transformer module in extracting information from local neighbor points. The corresponding segmentation results are shown in Fig.~\ref{f6}. It can be observed that other algorithms display significant segmentation errors at the boundaries of various components in the lamp and headphone targets. In contrast, the proposed algorithm exhibits minimal segmentation errors. At the same time, the global similarity computed in our method achieves a high level of accuracy compared to other algorithms, as shown in Fig.~\ref{f7}. In other algorithms, high response weights are assigned not only to points belonging to the wing part, but also to interfering points in the tail part. In contrast, in the proposed algorithm, high response weights are assigned exclusively to points belonging to the wing part. Moreover, no segmentation error is detected in the wing section in our method.

\subsection{Ablation Study}

\begin{table}[t]
\tabcolsep 7pt
\caption{Ablation study on ShapeNetPart.}
\label{t3}
\centering
\resizebox{\columnwidth}{!}{
\begin{threeparttable}
\begin{tabular}{ccccccccc}
\toprule
\multirow{2}{*}{Models} & \multicolumn{2}{c}{\makecell{Local Geometric Transformer}} & \multicolumn{2}{c}{\makecell{Global Semantic Transformer}} & \multirow{2}{*}{Ins.mIoU} & \multirow{2}{*}{Cat.mIoU} \\
\cmidrule(lr){2-3} \cmidrule(lr){4-5}
& \makecell{~~~Dis.weight} & \makecell{Geo.weight} & ~~~Glo.similairty & Glo.mask& & \\
\midrule
A & $~~~\checkmark$ & & & & 85.1 & 84.2 \\
B & & $~\checkmark$ & & & 85.5 & 84.5 \\
C & $~~~\checkmark$ & $~\checkmark$ & & & 86.6 & 85.0 \\
D & $~~~\checkmark$ & $~\checkmark$ & $~~\checkmark$ & & 87.0 & 85.3 \\
E & $~~~\checkmark$ & $~\checkmark$ & & $\checkmark$ & 87.2 & 85.4 \\
F & $~~~\checkmark$ & $~\checkmark$ & $~~\checkmark$ & $\checkmark$ & \textbf{87.5} & \textbf{85.6} \\
\bottomrule
\end{tabular}
\begin{tablenotes}[flushleft]
\small
\item Note: Dis.weight: distance weight. Geo.weight: geometric weight. Glo.similarity: global similarity.
\item \hspace{7.8mm} Glo.mask: global mask.
\end{tablenotes}
\end{threeparttable}}
\end{table}

\begin{table}[t]
\centering
\begin{minipage}{\textwidth}
\tabcolsep 3.2pt
\begin{minipage}[t]{0.48\textwidth}
\centering
\makeatletter\def\@captype{table}
\caption{Effect of different neighbor size settings.}
\label{t4}
\begin{tabular}{ccccccc}
    \toprule
    \textit{k} & 8 & 16 & 24 & 32 & 48\\
    \midrule
    Ins.mIoU & 86.5 & 87.2 & \textbf{87.5} & 87.4 & 87.2  \\
    Cat.mIoU & 84.7 & 85.3 & \textbf{85.6} & 85.3 & 85.1  \\
    \bottomrule
\end{tabular}
\end{minipage}%
\hfill
\begin{minipage}[t]{0.48\textwidth}
\centering
\makeatletter\def\@captype{table}
\caption{Effect of different head number settings.}
\label{t5}
\begin{tabular}{ccccccc}
    \toprule
    Heads & 1 & 2 & 4 & 6 & 8\\
    \midrule
    Ins.mIoU. & 87.1 & 87.3 & \textbf{87.5} & 87.4 & 87.2  \\
    Cat.mIoU. & 85.2 & 85.4 & \textbf{85.6} & 85.5 & 85.3  \\
    \bottomrule
\end{tabular}
\end{minipage}
\end{minipage}
\end{table}

\begin{table}[t]
\tabcolsep 5pt
\caption{Investigation of different operators.}
\centering
\label{t6}
\resizebox{0.8\columnwidth}{!}{
\begin{tabular}{ccccc}
\toprule
Operators & Concat & Summation & Average & Hadamard product \\
\midrule
Cat.mIoU & 85.3 & 85.3 & 85.4 & \textbf{85.6} \\
Ins.mIoU & 87.0 & 87.1 & 87.2 & \textbf{87.5} \\
 \bottomrule
\end{tabular}%
}
\end{table}

\textbf{Effects of different components.} To further illustrate the validity of the transformer modules in our method, we design an ablation study on the ShapeNetPart dataset as shown in Table~\ref{t3}. It can be observed from model A and B that the improvement is minimal when replacing the distance weights with geometric weights alone. It suggests that while geometric weight diminishes the effect of neighbor points with geometric disparity to the query point, the model struggles to focus on meaningful neighbor points. This is evident since the weights assigned to neighbor points that share similar geometry with the query point are nearly equal. In contrast, model C shows a significant performance improvement. Since only neighbor points that share similar geometry and close to the query point are assigned large weights.

Furthermore, from the results of model D and E, it is evident that modeling global dependencies is crucial for point cloud segmentation. Also, we can see that solely employing the global mask generated by the multi-head voting strategy results in superior performance compared to solely using global similarity. At last, the optimal accuracy is achieved when the global mask is combined with the global similarity. Since the weight information belonging to interfering points is filtered out in the refined similarity.

\noindent\textbf{Effect of different neighborhood size.} The number of neighbor points becomes a crucial parameter when employing the local geometric transformer module. It determines the size of the receptive field for the local point cloud region. We test our model on the ShapeNetPart benchmark with various settings to ascertain the optimal value.  As shown in Table~\ref{t4}, the performance improves as the parameters $k$ increase. This suggests that expanding the receptive field by considering more neighbor points enhances the model's ability to capture relevant features and contexts. However, further increasing the value of $k$ may cause the performance of the model to degrade. Because it may introduce some geometrically similar but irrelevant point information, which could impact the local geometric transformer module's ability to extract valid features.

\noindent\textbf{Effects of different head number.} We test our model on ShapeNetPart to evaluate the impact of different head number settings on the model performance. The relevant results are shown in Table~\ref{t5}. When the number of attention heads is set to one, the multi-head voting strategy fails to reduce the weight of interfering points. As a result, interference points may still exhibit high response weight in refined similarity. Meanwhile, As the number of heads increases, the model's performance gradually improves. Since multiple heads aid the model to preserve meaningful global information. However, the performance deteriorates when the number of heads become larger.

\noindent\textbf{Effects of different operator.} We employ different operators that integrate geometric weights within the local geometric transformer module to evaluate their performance. The results are presented in Table~\ref{t6}. Concat, Summation, Average, and Hadamard product denote the element-wise operations of concatenating over the channel, adding, averaging, and multiplying the geometric weight and distance weight, respectively. As can be seen, four operations have relatively minor effects on the final performance. But the Hadamard product obtains the best results. Since the Hadamard product achieves a substantial reduction in the weights of neighbor points with geometrical disparity. Simultaneously, it preserves the weights of points within the same class neighborhood, effectively balancing the influence of various data points.

\section{Conclusion}
In this paper, to better leverage local geometric information and accurately capture long-range semantic relationships within the transformer framework, we propose a novel transformer network named GSTran for point cloud segmentation. GSTran mainly consists of two essential modules: a local geometric transformer and a global semantic transformer. In the local geometric transformer module, we explicitly compute the geometric disparity. This allows us to amplify the affinity with geometrically similar neighbors and simultaneously suppress the association with other neighbors. In the global semantic transformer module, we design a multi-head voting strategy. This strategy computes the semantic similarity for each point over a global spatial range, capturing more accurate semantic information. Experiments with competitive performance on public datasets and further analysis demonstrate the effectiveness of our method.

\bibliographystyle{splncs04}
\bibliography{Reference}


\section*{Appendix}

Experiments on robustness, outdoor point cloud scenarios, and others are included in the supplementary material. For more details, please refer to the links below. Either of the following two links can be chosen.\\

\noindent\textcolor{black}{\href{https://drive.google.com/file/d/1rS36mBizZS4yHw4tcuOc5JAYDYLr1SUk/view?usp=sharing}{\underline{\textbf{Link 1: Google Drive.}}}}\\
\textcolor{black}{\text{https://drive.google.com/file/d/1rS36mBizZS4yHw4tcuOc5JAYDYLr1SUk/vie}}\\
\textcolor{black}{\text{w?usp=sharing}}

\noindent\textcolor{black}{\href{https://pan.baidu.com/s/1T3hOOrgMKvwmQOvGTzaeVQ?pwd=1234}{\underline{\textbf{Link 2: Baidu Drive. Password: 1234}}}}\\
\textcolor{black}{
\text{https://pan.baidu.com/s/1T3hOOrgMKvwmQOvGTzaeVQ}} \\
\textcolor{black}{\text{Password: 1234}}

\end{document}